\documentclass{article}

\usepackage[preprint]{neurips_2023}

\usepackage[utf8]{inputenc} 
\usepackage[T1]{fontenc}    
\usepackage{hyperref}       
\usepackage{url}            
\usepackage{booktabs}       
\usepackage{amsfonts}       
\usepackage{nicefrac}       
\usepackage{microtype}      
\usepackage{xcolor}         
\usepackage{enumitem}
\usepackage{amsmath}

\usepackage{amssymb}
\usepackage{bbm}
\usepackage{mathtools}
\usepackage{multirow}
\usepackage{subfigure}
\usepackage{amsthm}
\usepackage{bm}
\usepackage{adjustbox}
\usepackage{wrapfig}
\usepackage{algpseudocode}
\usepackage{graphicx}
\theoremstyle{plain}
\newtheorem{theorem}{Theorem}
\newtheorem*{theorem1}{Theorem 1}

\theoremstyle{definition}
\newtheorem{definition}[theorem]{Definition}

\theoremstyle{remark}

\DeclareMathOperator*{\argmax}{arg\,max}

\usepackage{titlesec}
\titlespacing*{\section}
{0pt}{2mm}{2mm}
\titlespacing*{\subsection}
{0pt}{1.5mm}{1.5mm}
\titlespacing*{\subsubsection}
{0pt}{1mm}{1mm}

\newcommand{\bz}{{\bf z}}
\newcommand{\bh}{{\bf h}}

\newcommand{\cG}{{\mathcal G}}
\newcommand{\cT}{{\mathcal T}}
\newcommand{\cS}{{\mathcal S}}
\newcommand{\cX}{{\mathcal X}}
\newcommand{\cY}{{\mathcal Y}}
\newcommand{\cV}{{\mathcal V}}
\usepackage{lipsum}
\usepackage[linesnumbered,ruled,vlined]{algorithm2e}

\SetCommentSty{mycommfont}

\SetKwInput{KwInput}{Input}                
\SetKwInput{KwOutput}{Output}              

\title{Domain Generalization Deep Graph Transformation}

\author{%
  Shiyu Wang \\
  Department of Biostatistics and Bioinformatics\\
  Emory University\\
  \texttt{shiyu.wang@emory.edu} \\
  \And
  Guangji Bai \\
  Department of Computer Science \\
  Emory University \\
  \texttt{guangji.bai@emory.edu} \\
  \AND
  Qingyang Zhu \\
  Department of Environmental Health \\
  Emory University \\
  \texttt{qingyang.zhu@emory.edu} \\
  \And
  Zhaohui Qin \\
  Department of Biostatistics and Bioinformatics \\
  Emory University \\
  \texttt{zhaohui.qin@emory.edu} \\
  \And
  Liang Zhao\\
  Department of Computer Science \\
  Emory University \\
  \texttt{liang.zhao@emory.edu} \\
}

\begin{document}

\maketitle

\begin{abstract}
Graph transformation that predicts graph transition from one mode to another is an important and common problem. Despite much progress in developing advanced graph transformation techniques in recent years, the fundamental assumption typically required in machine-learning models that the testing and training data preserve the same distribution does not always hold. As a result, domain generalization graph transformation that predicts graphs not available in the training data is under-explored, with multiple key challenges to be addressed including (1) the extreme space complexity when training on all input-output mode combinations, (2) difference of graph topologies between the input and the output modes, and (3) how to generalize the model to (unseen) target domains that are not in the training data. To fill the gap, we propose a multi-input, multi-output, hypernetwork-based graph neural network (MultiHyperGNN) that employs a encoder and a decoder to encode topologies of both input and output modes and semi-supervised link prediction to enhance the graph transformation task. Instead of training on all mode combinations, MultiHyperGNN preserves a constant space complexity with the encoder and the decoder produced by two novel hypernetworks. Comprehensive experiments show that MultiHyperGNN has a superior performance than competing models in both prediction and domain generalization tasks. The code of MultiHyperGNN is in \url{https://github.com/shi-yu-wang/MultiHyperGNN}.
\end{abstract}
\section{Introduction}
Graph is a ubiquitous data structure characterized by node attributes and the graph topology that describe objects and their relationships. Many tasks on graphs ask for predicting a graph (i.e., graph topology or node attributes) from another one. Applications of such graph transformation include traffic forecasting between two time stamps based on traffic flow~\citep{li2017diffusion, yu2017spatio}, fraud detection between transactional periods~\citep{van2022inductive}, and chemical reaction prediction according to molecular structures~\citep{guo2019deep, pan2022property, wang2022multi}. 

\begin{figure}[h]
\begin{center}
\includegraphics[width=\textwidth]{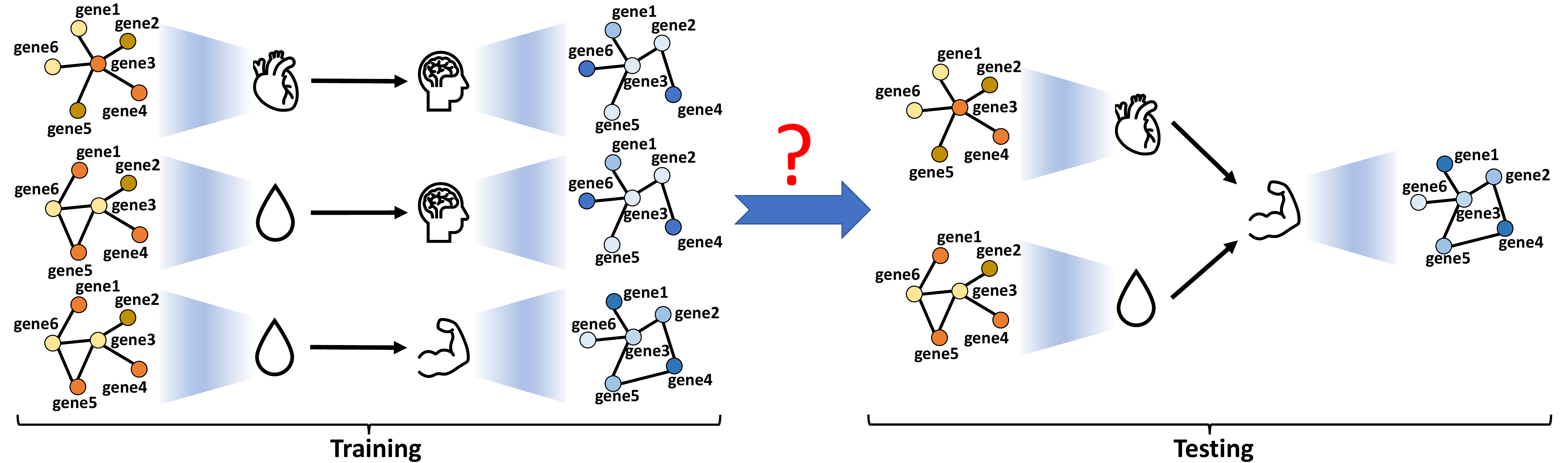}
\caption{Graph transformation predicts gene expression of the output tissue (e.g., brain and muscle) from gene expression of the input tissue (e.g., heart and whole blood). The yellow and the blue network represents the gene-gene network in the input tissue and the output tissue, respectively. Node colors indicate different expression values. During training we train the model on input-target tissue pairs, whereas during testing we generalize the model to unseen mode pairs or even unseen modes.}
\label{fig:gene}
\end{center}
\end{figure}

Despite of a wide spectrum of applications, graph transformation still faces major issues such as insufficient samples of graph pairs for training the model. For instance, as shown in Figure~\ref{fig:gene}, if the model is trained to predict gene-gene network on specific tissue pairs (e.g., from heart and blood to brain, from blood to muscle), but in testing process, one may want to generalize the model to unseen tissue pairs (e.g., from heart to muscle) or even to tissues unavailable in the training data. If so, the performance of the graph transformation model may deteriorate due to domain distribution gaps~\citep{quinonero2008dataset}. Therefore, it is imperative and crucial to improve the generalization ability of graph transformation models to generalize the learned graph transformation to other (unseen) graph transformations, namely domain generalization graph transformation.

Domain generalization graph transformation, nevertheless, is still under-explored by the machine-learning community due to the following challenges: \textbf{(1) High complexity in the training process}. To learn the distribution of graph (or mode) pairs in training data, we need to learn the model by traversing on all combinations of input modes to predict all combinations of output modes. In this case, the training complexity would be exponential if we train a single model for all possible input-output mode combinations; \textbf{(2) Graph transformation between topologically different modes}. The existing works regarding graph transformation predict node attributes conditioning on either the same topology or the same set of nodes of input and output modes~\citep{battaglia2016interaction, yu2017spatio, guo2019deep}. Performing graph transformation across modes with varying topologies, including different edges and even varying graph sizes, is a difficult task. Main challenges include how to learn the mapping between distinct topologies and how to incorporate the topology of each mode to enhance the prediction task; \textbf{(3) Learning graph transformation involving unseen domains and lack of training data.} Graph transformation usually requires both the source and target domains to be visible and have adequate training data to train a sophisticated model. However, during the prediction phase, we may be asked to predict a graph in an unseen target domain. Learning such transformation mapping without any training data is an exceedingly challenging task.

To fill the gap, we propose a novel framework for domain generalization graph transformation via a multi-input, multi-output hypernetwork-based graph neural networks (MultiHyperGNN). Our contributions are summarized as follows:
\begin{itemize}[leftmargin=*]
    \item \textbf{A novel multi-input, multi-output framework of graph transformation is proposed.} We aim at graph transformation for predicting node attributes across multiple input and output modes, by introducing a novel framework based on the multi-input, multi-output training strategy. The space complexity of the model is reduced from exponential to constant in training process.
    \item \textbf{An encoder and a decoder are developed for graph transformation between topologically different modes.} 
    To achieve the graph transformation between topologically different modes, MultiHyperGNN has an encoder and a decoder to encode the graph in the input and output mode, respectively. Additionally, MultiHyperGNN performs semi-supervised link prediction to complete the output graph, enabling the model to generalize to all nodes in the output mode.
    \item \textbf{Two hypernetworks are used to produce the encoder and the decoder for domain generalization.} We design two novel hypernetworks that produce the encoder and the decoder. Mode-specific meta information serves as the input to guide the hypernetwork to produce the corresponding encoder or decoder, and generalize to unseen target domains. 
    \item \textbf{The performance of MultiHyperGNN is experimentally superior}. We conduct extensive experiments to demonstrate the effectiveness of MultiHyperGNN on two real-world datasets. The experimental results show that MultiHyperGNN is superior than competing models.
\end{itemize}
This paper introduces the existing works on domain generalization graph transformation in Section~\ref{sec:rw}. Next, the problem is formally defined in Section~\ref{sec:prob}. Details of the proposed model, MultiHyperGNN, is discussed in Section~\ref{sec:model}, followed by the experiments in Section~\ref{sec:expr} and conclusion in Section~\ref{sec:con}.
\section{Related works}
\label{sec:rw}
\subsection{Graph transformation}
Graph transformation maps graph from one mode to another~\citep{du2021graphgt}. Some of the existing works predict node attributes given fixed graph topology. Li et al.~\cite{li2017diffusion} predicted traffic forecasting by incorporating both spatial and temporal dependency. Battaglia et al.~\cite{battaglia2016interaction} predicted velocities of objects on the subsequent time step. Some works instead predict graph topology. Guo et al.~\cite{guo2022deep} learned the global and local translation with graph convolution and deconvolution layers. Other works instead simultaneously predict node attributes and graph topology. Guo et al.~\cite{guo2019deep} solved node-edge joint translation with a multi-block network. Lin et al.~\citep{lin2020meta} applied graph attention to the co-evolution of node and edge states. When predicting node attributes, nevertheless, the assumption of fixed graph topology in both input and output modes may not always hold. Graph transformation that can handle topologies of both modes remains to be explored.
\subsection{Domain generalization}
Machine learning systems usually assume the same distribution between the training and the testing data, whereas generalizing trained models to unseen data is significant in fields such as semantic segmentation~\citep{gong2019dlow, dou2019domain}, fault diagnosis~\citep{li2020domain, zheng2020deep}, natural language processing~\citep{wang2020meta, garg2021learn}, etc~\citep{du2021adarnn, qian2021latent}. Du et al.~\cite{du2021adarnn} applied domain generalization to time series modeling by an RNN-based model to solve the temporal covariate shift. Qian et al.~\citep{qian2021latent} applied domain generalization to sensor-based human activity recognition by learning the domain-invariant modules to disentangle different persons. Gong et al.~\citep{gong2019dlow} translated images from the input to the output mode while producing a sequence of intermediate modes for domain generalization. Wang et al.~\citep{wang2020meta} used meta learning that targets zero-shot domain generalization for semantic parsing. Chen et al.~\citep{chen2022compound} assumed that the domain label is unavailable for training and the model needs to identify latent domain structure and their semantic correlations, which may expect sufficient expressiveness of the representation learning process. 
\subsection{Hypernetworks}
A hypernetwork is a neural network that generates weights of another neural network~\citep{ha2016hypernetworks}. Hypernetwork has a broad spectrum of applications, including image classification tasks~\citep{sun2017hypernetworks, sendera2023hypershot}, image editing~\citep{alaluf2022hyperstyle}, robotic control~\citep{huang2021continual, rezaei2022hypernetworks} and language models~\citep{volk2022example, zhang2022hyperpelt}. Noticeably, hypernetwork has also been employed for domain generalization. Qu et al.~\citep{qu2022hmoe} used hypernetworks to generate weights of experts while allowing experts to share meta-knowledge. This model needs to generate multiple classifiers and take the weighted sum for the final prediction, whose training space complexity is linear to the number of classifiers. Sendera et al.~\citep{sendera2023hypershot} proposed HyperShot where the kernel-based representation of the support examples is fed to hypernetwork to create the classifier for few-shot learning. Bai et al.~\cite{bai2022temporal} utilized hypernetworks to produce graph classifiers, but with only time stamps as the input of hypernetworks to focus on temporal domain generalization. Despite of the wide use of hypernetworks for domain generalization, the research of hypernetworks to generate GNNs is limited. In our work, we design two novel hypernetworks to guide the domain generalization on graph transformation tasks.
\section{Problem formulation}
\label{sec:prob}
Suppose we have $N$ modes of graphs composed of $p$ nodes: $\mathcal G=\{\mathcal G^{(1)}, \mathcal G^{(2)}, ..., \mathcal G^{(N)}\}$, where each mode contains graphs with the same topology. Specifically, suppose there are $n$ independent samples in the dataset, and for each sample $i$ in the mode $j$, denote $G_{i}^{(j)}=\{A^{(j)}, X_{i}^{(j)}\}\in\mathcal G^{(j)}$, where $A^{(j)}\in\mathbb{R}^{p_j\times p_j}$ is the graph of size $p_j\le p$ and $X_{i}^{(j)}\in\mathbb R^{p_j\times d}$ is the node attributes with $d$ features. Note that the graph of $\mathcal G^{(j)}$ may not contain all $p$ nodes in mode $j$, all other nodes are disjointed with each other and with $p_j$ nodes in $\mathcal G^{(j)}$. We further assume each mode $j$ can be characterized by its meta information $m^{(j)}$. For instance, the mode $\mathcal G^{(j)}$ can be a specific human tissue $j$ that has the gene-gene expression network $G_i^{(j)}$ for a patient $i$. There are $p$ human genes expressed in various human tissues but $G_i^{(j)}$ only contains $p_j$ of them.

Next, we formally formulate the task \textbf{domain generalization graph transformation} as below:
\begin{definition}[Domain generalization graph transformation]
Let $\mathcal S=\{\mathcal X\times\mathcal Y: \mathcal X\in\mathcal P(\mathcal G), \mathcal Y\in\mathcal P(\mathcal G-\mathcal X)\}$ be the source domain where we train the graph-transformation model $f:\mathcal X\rightarrow \mathcal Y$, which predicts node attributes in $\mathcal Y$ from node attributes in $\mathcal X$. $\mathcal P(\cdot)$ is the power set excluding the empty set. \textit{Domain generalization graph transformation} learns the $f$ so that the prediction error on $f: \mathcal X^{\mathcal T}\rightarrow\mathcal Y^{\mathcal T}$ is minimized, where $\mathcal X^{\mathcal T}\times \mathcal Y^{\mathcal T}$ is the target domain s.t. $\mathcal X^{\mathcal T}\times \mathcal Y^{\mathcal T}\notin \mathcal S$.
\label{def:dggt}
\end{definition}
Domain generalization graph transformation is exceptionally difficult due to the following challenges:

\textbf{Challenge 1: High complexity in training process.} For training the graph transformation $f:\mathcal{X}\rightarrow\mathcal{Y}$, where $\mathcal X\times\mathcal Y\in\mathcal S$, conventionally we need to train $O(3^N)$ models to handle all possible mode combinations in $\mathcal S$, which is rather computationally intensive. 

\textbf{Challenge 2: Topological difference between input and output domains.} When the input mode and the output mode have different topologies, how to utilize topologies of both modes to jointly contribute to graph transformation remains to be explored. An intuitive way is to employ two graph encoders to respectively encode the graph topology of both modes, but how to form the graph-transformation model on $\mathcal S$ with only two such encoders is still challenging.

\textbf{Challenge 3: Generalization to unseen domains.} Even if it is possible that we train the model on all combinations of modes in $\mathcal S$, how to learn the graph transformation that can efficiently predict the graph in an unseen target domain is still challenging.
\section{Domain generalization deep graph transformation}
\label{sec:model}
\subsection{Overview of MultiHyperGNN}
\label{sec:overview}
\begin{figure}[h]
\begin{center}
\includegraphics[width=0.6\textwidth]{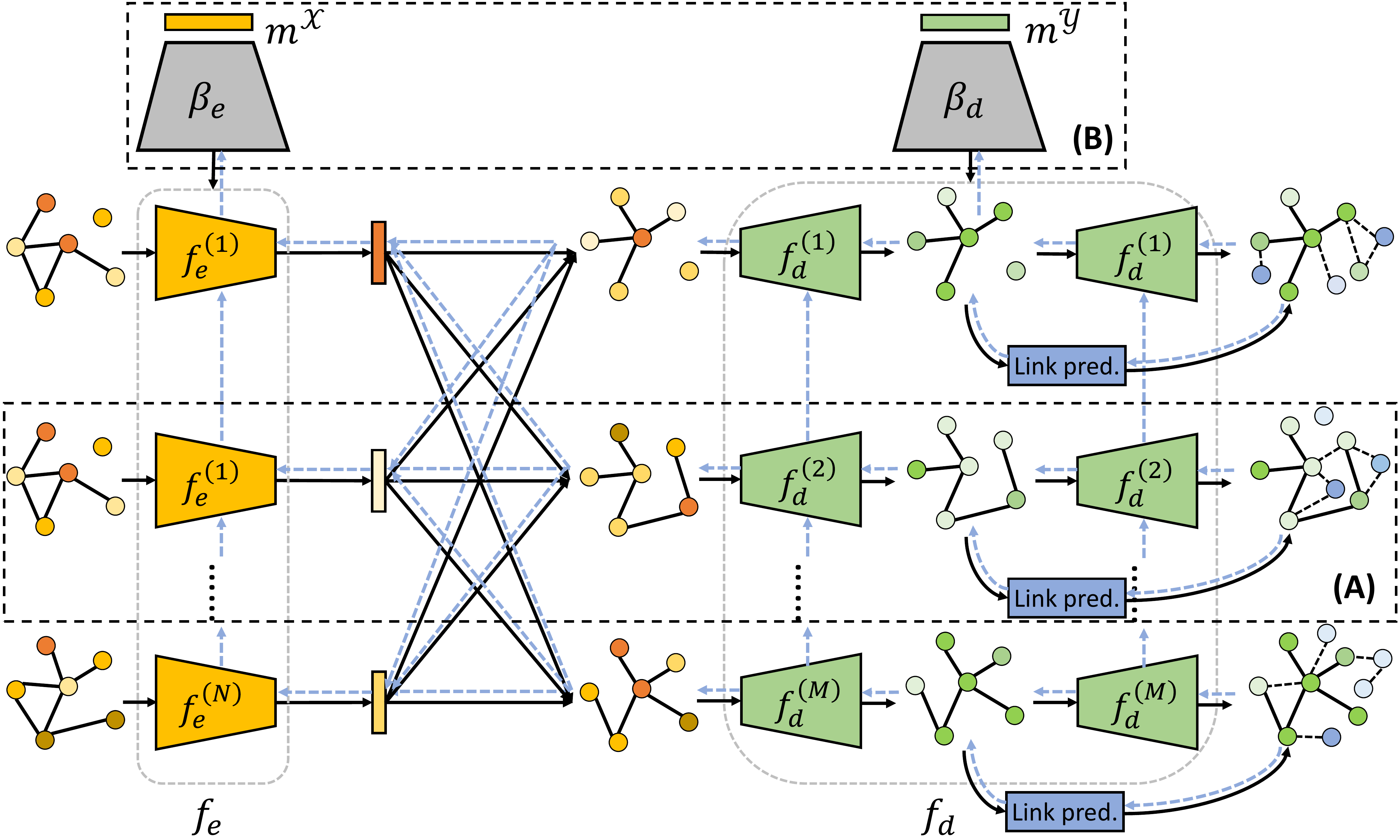}
\caption{Overview of MultiHyperGNN for the graph transformation from $N$ input to $M$ output modes. $\{f_e^{(j)}\}_{j\in\mathcal{X}}$ are encoders generated by the encoder hypernetwork $\beta_e$. $\{f_d^{(k)}\}_{k\in\mathcal{Y}}$ are decoders generated by the decoder hypernetwork $\beta_d$. $m^{\mathcal{X}}$ and $m^{\mathcal{Y}}$ are mode-specific meta information. "Link pred." is semi-supervised link prediction. Blue dotted line is the flow of back-propagation.}
\label{fig:hypergnn}
\end{center}
\end{figure}

For the first challenge, instead of including all the exponentially many modes by separately training all their combinations, we collectively train all the modes together to avoid the duplication of modes and reduce the time complexity to linearity (Figure~\ref{fig:hypergnn}). The details are given in this section. To address the heterogeneity of node set and topology, we propose a novel encoder-decoder framework in Section~\ref{sec:graphtrans} (Figure~\ref{fig:hypergnn} (A)). Moreover, each input mode requires an encoder while each output mode needs a decoder, which can be any type of GNNs such as Graph Convolutional Network (GCN), Graph Isomorphism Network (GIN) and Graph Attention Network (GAT). To learn the encoder and the decoder for unseen modes, we propose to train two hypernetworks that can respectively generate any encoder or decoder given the meta information of the mode in Section~\ref{sec:hyper} (Figure~\ref{fig:hypergnn} (B)). Furthermore, we provide a theoretical assurance that an ample amount of meta-information will result in improved generalization accuracy when extrapolating to unexplored domains.

Let $j\in\mathcal{X}$ be the $j$-th mode in $\mathcal{X}$ and $k\in\mathcal Y$ be the $k$-th mode in $\mathcal Y$, where $\mathcal X\in\mathcal P(\mathcal G)$ and $\mathcal Y\in\mathcal P(\mathcal G-\mathcal X)$ as in Def.~\ref{def:dggt}. As shown in Figure~\ref{fig:hypergnn} (A), to predict $X_i^{(k)}$, we employ encoders (i.e., $f_e=\{f_e^{(j)}\}_{j\in\mathcal{X}}$) and decoders (i.e., $f_d=\{f_d^{(k)}\}_{k\in\mathcal{Y}}$) to encode the topology of both input and output modes:
\small
\begin{eqnarray}
\hat X_i^{(k)}=f_d^{(k)}(A^{(k)}, \sigma_1(\{f_e^{(j)}(G_i^{(j)}; \beta_e^{(j)})\}_{j\in \mathcal X})); \beta_d^{(k)})), \ \ k\in\mathcal Y
 \label{eq:f}
\end{eqnarray}
\normalsize
Namely, to predict node attributes of any $k\in\mathcal Y$ from modes in $\mathcal X$, we first encode topologies of modes in $\mathcal X$ via $f_e^{(j)}(G_i^{(j)}; \beta_e^{(j)})$, $j\in\mathcal X$. Then we aggregate embeddings of all these modes via the pooling function $\sigma_1(\cdot)$ and feed it with the graph topology $A^{(k)}$ into $f_d^{(k)}$ to predict $X_i^{(k)}$, where $k\in\mathcal Y$. To reduce the heavy complexity of the training process due to the exponential number of choices of $\mathcal X\times\mathcal Y\in\mathcal S$ (Def.~\ref{def:dggt}) and generalize the mode to unseen domains, instead of training separately for each $\mathcal X\times\mathcal Y$, as shown in Figure~\ref{fig:hypergnn} (B), we borrow two hypernetworks (i.e., $\beta_e$ and $\beta_d$) to produce all encoders and decoders with the corresponding mode-specific meta information:
\small
\begin{eqnarray}
    \beta_e^{(j)}=\beta_e(m^{(j)};\gamma_e),\ \ j\in \mathcal{X}; \ \ \beta_d^{(k)}=\beta_d(m^{(k)};\gamma_d),\ \ k\in \mathcal{Y},
    \label{eq:hyper}
\end{eqnarray}
\normalsize
where $\gamma_e$ and $\gamma_d$ parameterize $\beta_e$ and $\beta_d$, respectively, and are learned during training process. Therefore, Eq.~\ref{eq:f} is re-parameterized by $\gamma_{\mathcal{X}\rightarrow\mathcal{Y}}=\{\gamma_e, \gamma_d\}$:
\small
\begin{eqnarray}
\hat{X}_i^{(k)} &=& f_d^{(k)}(A^{(k)}, \sigma_1(\{f_e^{(j)}(G_i^{(j)};\beta_e(m^{(j)};\gamma_e))\}_{j\in\mathcal X}); \beta_d(m^{(k)};\gamma_d)), \ \ k\in\mathcal Y\nonumber \\
&=&f_{\gamma_{\mathcal{X}\rightarrow\mathcal{Y}}}(A^{(k)}, \{G_i^{(j)}, m^{(j)}\}_{j\in\mathcal X};\gamma_{\mathcal{X}\rightarrow\mathcal{Y}}), \ \ k\in\mathcal Y,
 \label{eq:pred}
\end{eqnarray}
\normalsize
where $f_{\gamma_{\mathcal{X}\rightarrow\mathcal{Y}}}=\{f_d^{(k)}*\{f_e^{(j)}\}_{j\in\mathcal X}\}_{k\in\mathcal{Y}}: \mathbb{R}^{N\times p} \rightarrow \mathbb{R}^{M\times p}$ formularizes the graph transformation that predicts node attributes of
$M$ modes in $\mathcal Y$ from $N$ modes in $\mathcal X$. 

As long as $\hat{X}_i^{(k)}$ is predicted via Eq.~\ref{eq:pred}, we mathematically formulate the first term of the learning objective of MultiHyperGNN as follows:
\small
\begin{eqnarray}
    \mathcal L_1(\gamma_e, \gamma_d)=\sum_{i=1}^{n}\ell(\{\hat{X}_i^{(k)}\}_{k\in\mathcal{Y}}, \{X_i^{(k)}\}_{k\in\mathcal{Y}})
    \label{eq:obj}
\end{eqnarray}
\normalsize
where $\ell(\cdot)$ measures the prediction error of $f_{\gamma_{\mathcal{X}\rightarrow\mathcal{Y}}}$ of each sample, such as mean squared error (MSE), mean absolute error (MAE), etc. $n$ is the total number of samples in training data.

Since the size of the source domain $\mathcal S$ is $O(3^N)$, leading to an exponential space complexity of $O(3^N)$ with the space of trainable parameters as $\{\mathcal P(\{\beta_e^{(j)}\}_{j\in\mathcal X})\times \mathcal P(\{\beta_d^{(k)}\}_{k\in\mathcal Y}): \mathcal X\in\mathcal P(\mathcal G), \mathcal Y\in\mathcal P(\mathcal G-\mathcal X)\}$. MultihyperGNN reduces the space complexity to $O(1)$. 
\subsection{Graph transformation on topologically different domains}
\label{sec:graphtrans}
Traditional graph-transformation models encounter significant challenges when attempting to handle modes with different graph topologies (i.e., $A^{(j)}\ne A^{(k)}, p_j\ne p_k$). To address this issue, as shown in Figure~\ref{fig:hypergnn} (A), we propose GNN-based encoder $f_e^{(j)}$ and decoder $f_d^{(k)}$ that encode the graph of modes $j\in\mathcal{X}$ and $k\in\mathcal{Y}$, perform semi-supervised link prediction to complete the topology of the output mode $\mathcal G^{(k)}$ and enable the model to predict all $p$ nodes. Let $\cV^{(j)}$ and $\cV^{(k)}$ be sets of nodes contained in the graph of modes $j$ and $k$, respectively, and $|\cV^{(j)}|=p_j$, $|\cV^{(k)}|=p_k$. Since $p_j\ne p_k$, to match the input dimension of $f_e^{(j)}$ and $f_d^{(k)}$, we expand $\mathcal G^{(j)}$ and $\mathcal G^{(k)}$ by the union of their nodes and obtain $\Tilde{\mathcal{G}}^{(j)}$ and $\Tilde{\mathcal{G}}^{(k)}$ with node sets: $\Tilde \cV^{(j)}=\Tilde{\cV}^{(k)}=\cV^{(j)}\bigcup \cV^{(k)}$, and $|\Tilde \cV^{(j)}|=\Tilde p_j=|\Tilde \cV^{(k)}|=\Tilde p_k\le p$. Those newly added nodes are self-connected and are disjointed with other nodes.
\subsubsection{Encoder}
\label{sec:se}
For the $i$-th sample, the encoder $f_e^{(j)}$ encodes the topology and node attributes of the mode $j$ into the latent embedding $\bz_i^{(j)}\in\mathbb{R}^{l}$, where $l$ is the hidden dimension:
\small
\begin{eqnarray}
\bz_i^{(j)}=f_e^{(j)}(\Tilde{G}_i^{(j)}; \beta_e^{(j)})=\text{GNN}(\Tilde{G}_i^{(j)}; \beta_e^{(j)}).
\label{eq:se}
\end{eqnarray}
\normalsize
Based on Eq.~\ref{eq:hyper}, the encoder $f_e^{(j)}$ is generated by the hypernetwork $\beta_e$ guided by the mode-specific meta information $m^{(j)}$. Therefore, Eq.~\ref{eq:se} becomes $\bz_i^{(j)}=\text{GNN}(\Tilde{G}_i^{(j)}; \beta_e(m^{(j)};\gamma_e))$, where $\gamma_e$ is mode-invariant and parameterizes all encoders $\{f_e^{(j)}\}_{j\in\mathcal X}$.

\subsubsection{Decoder}
\label{sec:td}
Once $\{\bz_i^{(j)}\}_{j\in\mathcal X}$ is obtained for all modes in $\mathcal X$, we apply the decoder $f_{d}^{(k)}$ that decodes $\{\bz_i^{(j)}\}_{j\in\mathcal X}$ and encodes the topology $\Tilde A^{(k)}$ of the output mode $k\in\mathcal Y$ to predict node attributes of $\Tilde \cV^{(k)}$:
\small
\begin{eqnarray}
    \hat{\Tilde{X}}_i^{(k)}=f_d^{(k)}(\Tilde A^{(k)}, \sigma_1(\{\bz_i^{(j)}\}_{j\in\mathcal X}); \beta_d^{(k)})=\text{MLP}(\text{GNN}(\Tilde A^{(k)}, \sigma_1(\{\bz_i^{(j)}\}_{j\in\mathcal X}); \beta_{d, \text{GNN}}^{(k)}); \beta_{d, \text{MLP}}^{(k)}),
    \label{eq:td}
\end{eqnarray}
\normalsize
where the Multilayer Perceptron (MLP) serves as the prediction layer and $\beta_d^{(k)}=\{\beta_{d, \text{GNN}}^{(k)}, \beta_{d, \text{MLP}}^{(k)}\}$, generated by the hypernetwork $\beta_d$ with mode-specific meta information $m^{(k)}$. Then Eq.~\ref{eq:td} becomes:
\small
\begin{eqnarray}
    \hat{\Tilde{X}}_i^{(k)}=\text{MLP}(\text{GNN}(\Tilde A^{(k)}, \sigma_1(\{\bz_i^{(j)}\}_{j\in\mathcal X}); \beta_d(m^{(k)};\gamma_d)); \beta_d(m^{(k)};\gamma_d)),
    \label{eq:hypertd}
\end{eqnarray}
\normalsize
where $\gamma_d$ is model-invariant and parameterizes all target decoders $\{f_d^{(k)}\}_{k\in\mathcal Y}$. We further define $\mathcal V$ as the set of all nodes contained in $\mathcal G$ so that $|\mathcal V|=p$. Since $p\ge \Tilde p_j = \Tilde p_k$, now we have only predicted node attributes of $\Tilde \cV^{(k)}$, and the attributes of the remaining nodes $\mathcal V\setminus \Tilde \cV^{(k)}$ still need to be predicted.
\subsubsection{Semi-supervised link prediction}
We adopt the semi-supervised link prediction to complete the topology of the mode $k$ using graph auto-encoder~\citep{kipf2016variational} under the supervision of $\Tilde{A}^{(k)}$:
\small
\begin{eqnarray}
    \bh_i^{(k)} = \text{GNN}(\Tilde A^{(k)}, \hat{\Tilde X}_i^{(k)};\phi), \quad \hat A^{(k)} =\text{Sigmoid}(\bh_i^{(k)}\cdot (\bh_i^{(k)})^T)
    \label{eq:semi}
\end{eqnarray}
\normalsize
Then we compute the Binary Cross Entropy (BCE) between $\Tilde A^{(k)}$ and $\hat A^{(k)}$ as the second term of the learning objective:
\small
\begin{eqnarray}
    \mathcal L_2(\phi) = \sum_{k\in\mathcal Y}\text{BCE}(\Tilde A^{(k)}, \hat A^{(k)})=-\sum_{k\in\mathcal Y}\sum_{s=1}^{p_k}\sum_{t=1}^{p_k}\{\Tilde A_{st}^{(k)}\log \hat A_{st}^{(k)}+(1-\Tilde A_{st}^{(k)})\log (1-\hat A_{st}^{(k)})\}
    \label{eq:lpobj}
\end{eqnarray}
\normalsize
Once Eq.~\ref{eq:semi} is trained and $\hat \phi$ is learned, we perform link prediction and update $\Tilde A^{(k)}$ as follows:
\small
\begin{eqnarray}
    \bar \bh_i^{(k)} = \text{GNN}(\bar{A}^{(k)}, \bar{X}_i^{(j\rightarrow k)}; \hat \phi), \ \ \Tilde A^{(k)} \leftarrow \text{Sigmoid}(\bar \bh^{(k)}\cdot (\bar \bh^{(k)})^T),
    \label{eq:lp}
\end{eqnarray}
\normalsize
where $\bar{A}^{(k)}$ is the diagonal block matrix with $\Tilde A^{(k)}$ and the identity matrix $I\in\mathbb{R}^{(p-\Tilde p_k)\times (p-\Tilde p_k)}$ as diagonal blocks. Since the node attributes of $\Tilde \cV^{(k)}$ has been predicted in Eq.~\ref{eq:hypertd}, we only need to impute the attributes of $\mathcal{V}\setminus\Tilde{\mathcal{V}}^{(k)}$ with the corresponding attributes of modes in $\mathcal X$ as the input of GNN$(\cdot)$. Therefore, $\bar X_i^{(j\rightarrow k)}=[\hat{\Tilde X}_i^{(k)}, \sigma_2(\{X_{i}^{(j)}[\Tilde p_k:]\}_{j\in\mathcal X})]$ is the concatenation of previously predicted attributes $\hat{\Tilde X}^{(k)}$ (Eq.~\ref{eq:hypertd}) and the aggregated attributes of $\mathcal{V}\setminus\Tilde{\mathcal{V}}^{(k)}$ in modes $j\in\mathcal X$ via the pooling function $\sigma_2(\cdot)$. $\sigma_2(\cdot)$ is the mean pooling function across modes in $\mathcal X$ in implementation.

Once $\Tilde A^{(k)}$ is updated by Eq.~\ref{eq:lp}, we apply the decoder again in Eq.~\ref{eq:hypertd} to predict attributes of $\mathcal{V}\setminus\Tilde{\mathcal{V}}^{(k)}$:
\small
\begin{eqnarray}
    \hat{\bar X}_i^{(k)}=\text{MLP}(\text{GNN}(\Tilde A^{(k)}, \bar{X}_i^{(j\rightarrow k)}; \beta_d(m^{(k)};\gamma_d)); \beta_d(m^{(k)};\gamma_d))[\Tilde{p}_k:],
    \label{eq:predall}
\end{eqnarray}
\normalsize
Finally, the predicted node attributes in $k\in\mathcal Y$ are $\hat X_i^{(k)} = [\hat{\Tilde X}_i^{(k)}, \hat{\bar X}_i^{(k)}]$.
\subsection{Domain generalization via hypernetworks}
\label{sec:hyper}
In this section, we propose the encoder hypernetwork ($\beta_e$ in Figure~\ref{fig:hypergnn} (B)), the decoder hypernetwork ($\beta_d$ in Figure~\ref{fig:hypergnn} (B)), and the algorithm to learn them. The similarity among input and output modes is captured by meta information $m^{\mathcal{X}}=\{m^{(j)}\}_{j\in\mathcal{X}}$ and $m^{\mathcal{Y}}=\{m^{(k)}\}_{j\in\mathcal{Y}}$, respectively, which guide the encoder and the decoder hypernetwork to produce mode-specific encoders (i.e., $\{f_e^{(j)}\}_{j\in\mathcal{X}}$) and decoders (i.e., $\{f_d^{(k)}\}_{k\in\mathcal{Y}}$). When generalizing to unseen target domains, $\beta_e$ and $\beta_d$ can produce encoders and decoders of unseen modes given their meta information.
\subsubsection{Learning phase}
In the training process, we learn parameters $\gamma_e$, $\gamma_d$ of the encoder hypernetwork $\beta_e$ and the decoder hypernetwork $\beta_d$, respectively, on the source domain $\mathcal S=\{\mathcal X\times\mathcal Y: \mathcal X\in\mathcal P(\mathcal G), \mathcal Y\in\mathcal P(\mathcal G-\mathcal X)\}$. Specifically, we minimize the learning objective $\mathcal L$ of MultiHyperGNN:
\small
\begin{eqnarray}
    \mathcal L=\mathcal L_1(\gamma_e, \gamma_d) +\rho\cdot \mathcal L_2(\phi),
    \label{eq:overallobj}
\end{eqnarray}
\normalsize
where $\mathcal L_1$ and $\mathcal L_2$ are obtained from Eq.~\ref{eq:obj} and Eq.~\ref{eq:lpobj}, respectively, $\rho$ is the hyperparameter, and $\phi$ is another trainable paratemer for semi-supervised link prediction. In implementation, $\beta_e$ and $\beta_d$ are approximated by MLPs. The learning phase is also depicted in Algorithm~\ref{alg:learn}.
\begin{algorithm}[H]
\scriptsize
\DontPrintSemicolon
  \KwInput{$\{\mathcal {\Tilde G}^{(j)}, m^{(j)}\}_{j\in\mathcal X}$: topology, node attributes and meta information of source modes}
  \KwInput{$\{\Tilde A^{(k)}, m^{(k)}\}_{k\in\mathcal Y}$: topology and meta information of target modes}
  \KwInput{Initialized parameters $\gamma_e$, $\gamma_d$ and $\phi$}
  \KwOutput{$\{\hat X_i^{(k)}\}_{k\in\mathcal{Y}}$, i=1,2,...,n}
  \While{Converge}
   {
   \For{$k\in\mathcal Y$}{
   Compute attributes $\hat{\Tilde{X}}_i^{(k)}$ of $\Tilde{\mathcal{G}}^{(k)}$ via Eq.~\ref{eq:hypertd} for each sample $i$ in mode $k$\;
   Assemble $\bar A^{(k)}$ and $\bar X_i^{j\rightarrow k}$ as in Eq.~\ref{eq:lp}\;
   Perform link prediction and update $\Tilde{A}^{(k)}$ via Eq.~\ref{eq:lp}\;
   Compute attributes of the remaining nodes $\hat{\bar{X}}_i^{(k)}$ in $\mathcal{V}\setminus\Tilde{\mathcal{V}}^{(k)}$ via Eq.~\ref{eq:predall}\;
   Concatenate predicted attributes of all nodes: $\hat X_i^{(k)} = [\hat{\Tilde X}_i^{(k)}, \hat{\bar X}_i^{(k)}]$
   }
   Compute $\mathcal L_1(\gamma_e, \gamma_d)$ via Eq.~\ref{eq:obj}, $\mathcal L_2(\phi)$ via Eq.~\ref{eq:lpobj} and $\mathcal L$ via Eq.~\ref{eq:overallobj}\;
   Update $\gamma_e$, $\gamma_d$ and $\phi$ by stochastic gradient descent on $\mathcal L$
   }
\caption{Algorithm of learning phase}
\label{alg:learn}
\end{algorithm}
\subsubsection{Generalization phase}
Once $\hat{\gamma}_e$, $\hat{\gamma}_d$ are learned as parameters of $\beta_e$ and $\beta_d$, respectively, we generalize the model to the unseen target domain $\mathcal T=\mathcal X^{\mathcal T}\times \mathcal Y^{\mathcal T}$ by guiding $\beta_e$ and $\beta_d$ with the meta information of unseen modes $\{m^{(j)}\}_{j\in\mathcal X^{\mathcal T}}$ and $\{m^{(k)}\}_{k\in\mathcal Y^{\mathcal T}}$. Following Eq.~\ref{eq:f} and Eq.~\ref{eq:pred}, we have:
\small
\begin{eqnarray}
    \beta_e^{(j)} = \beta_e(m^{(j)}; \hat \gamma_e), \ j\in\mathcal X^{\mathcal T},\quad \beta_d^{(k)} = \beta_d(m^{(k)}; \hat \gamma_d), \ k\in\mathcal Y^{\mathcal T} \nonumber \\
    \hat{X}_i^{(k)} = f_d^{(k)}(A^{(k)}, \sigma(\{f_e^{(j)}(G_i^{(j)}; \beta_e^{(j)})\}_{j\in\mathcal Y^{\mathcal T}}); \beta_d^{(k)}), \ \ k\in\mathcal Y^{\mathcal T}
    \label{eq:dgf}
\end{eqnarray}
\normalsize
We theoretically prove that in the generalization phase our model can generalize to $\mathcal T$ given sufficient mode-specific meta information.
\begin{definition}[Generalization error]
Suppose $X_i^{(k)}=f_{\hat \gamma_{\cX\rightarrow\cY}}(A^{(k)}, \{G_i^{(j)}, m^{(j)}\}_{j\in\cX^\cT}; \hat \gamma_{\cX\rightarrow\cY})+\bm{\epsilon_i}$ following Eq.~\ref{eq:pred}, where $k\in\cY^\cT$ and $\hat{\gamma}_{\cX\rightarrow\cY}$ is estimated during training process on $\cS$. We define $\|\bm{\epsilon}_i\|_2^2$ as the \textit{generalization error} of sample $i$.
\label{def:ge}
\end{definition}
\begin{definition}[Sufficient meta information]
    We define $m^{(j)}$ and $m^{(k)}$ as the \textit{sufficient meta information} of the prediction $f_{\gamma_{\cX\rightarrow\cY}}$ if $j\in\cX$, $k\in\cY$, $m^{(j)}$ belongs to the space $\mathcal M$ that is a bijective mapping of the space of sufficient statistic of $\cG^{(j)}$, and $m^{(k)}=\argmax_{m} p(\cG^{(k)}\vert \cG^{(j)}, m)$.
\end{definition}
\begin{theorem1}
   For the mode $j\in\mathcal X^T$ and the mode $k\in\mathcal Y^T$, $\mathcal X^{\mathcal T}\times \mathcal Y^{\mathcal T}\in \mathcal T$, $m^{(j)}$ and $m^{(k)}$ are sufficient meta information of $j$ and $k$, compute $\hat X_i^{(k)}$ following Eq.~\ref{eq:pred} using $\mathcal G^{(j)}$, $A^{(k)}$, $m^{(j)}$ and $m^{(k)}$ and calculate the generalization error $\|\bm{\epsilon}_i\|_2^2$ as in Def.~\ref{def:ge}. Then compute $\{\hat{X}_i^{(k)'}\}_{k\in\mathcal Y^{\mathcal T}}$ following Eq.~\ref{eq:pred} using the same input but with $\forall m^{(j')}, j'\in\mathcal X$, $\forall m^{(k')}, k'\in\mathcal Y$ and $\mathcal X\times\mathcal Y\in\mathcal S$ as the input of $\beta_e$ and $\beta_d$. This leads to the generalization error $\|\bm{\epsilon}_i'\|_2^2$.  Assume $\bm\epsilon_i$ in Def.~\ref{def:ge} has a Gaussian distribution $\bm{\epsilon}_i\sim\mathcal N(\bm{0}, \bm{\sigma}^2)$, then we have $\|\bm{\epsilon}_i\|_2^2\le \|\bm{\epsilon}_i'\|_2^2$.
    \label{thm:1}
\end{theorem1}
The proof of the above theory is in Appendix A. Meta information is especially critical when producing encoders. The more informative it is, the more accurate the domain generalization is.
\section{Experiments}
\label{sec:expr}
This section reports the results of both quantitative and qualitative experiments that were performed to evaluate MultiHyperGNN and other competing models. 
\subsection{Dataset}
We conducted experiments on two real-world datasets: (1) \textbf{Genes}. We used gene expression data from Genotype-Tissue Expression Consortium~\citep{lonsdale2013genotype}, in which five tissues, whole blood (WB), lung (L), muscle skeletal (MS), sun-exposed skin (lower leg, LG), not-sun-exposed skin (suprapubic, S) were used and the gene-gene network was constructed by weighted correlation network analysis~\citep{langfelder2008wgcna} with the expression values as the node attributes. Meta information includes tissue type (lung, muscle, skin), location (trunk, leg, arm), structure (dense, rigid, spongy), function (movement, protection, gas exchange) and cell types (alveoli and bronchioles, cylindrical muscle fibers, epithelial cells); (2) \textbf{Climate}. We extracted data from the Goddard Earth Observing System Composition Forecasting across the US from 2019-2021. We collected the air temperature (T) for each state capital and then splitted a day into four modes: early morning (0:00AM-6:00AM), late morning (6:00AM-12:00PM), afternoon (12:00PM-18:00PM) and night (18:00PM-0:00AM). To construct the network, we used cities as graph nodes and air temperature in each city as node attributes. In each time period, two cities are connected if air temperatures between them have a high Pearson Correlation. We used the time period indicator (four-element, one-hot vector to indicate four periods) and various time stamps when collecting data as the meta information. Detailed introduction and summary statistics of datasets used are in Appendix B.

\begin{table*}[!tb]
\caption{Evaluation on prediction accuracy.}
\centering
\begin{adjustbox}{max width=0.9\textwidth}
\begin{tabular}{c c cc c cc c cc c cc c cc }\hline
\multirow{2}{*}{Model} && \multicolumn{2}{c}{Genes-L} && \multicolumn{2}{c}{Genes-LG} && \multicolumn{2}{c}{Genes-S} && \multicolumn{2}{c}{T-Afternoon} && \multicolumn{2}{c}{T-Night} \\\cline{3-4} \cline{6-7} \cline{9-10} \cline{12-13} \cline{15-16}
&& MSE & PCC && MSE & PCC && MSE & PCC && MSE & PCC && MSE & PCC \\ \hline\hline
ED-GNN &&1.9810  &0.6072  &&2.1289  &0.5795  &&2.1925  &0.5764 && 59.3010 & 0.4539 && 84.0824 & 0.4187  \\
MHM && 2.0126 & 0.5913 && 2.0153 & 0.5312 && 2.0384 & 0.5816&& 61.2798 & 0.4300 && 69.8599 & 0.4207  \\
IN && 2.0182 & 0.6026 && 2.2019 & 0.5683 && 2.1304 &0.5377 && 60.8755 & 0.4650 && 71.0456 &  0.4210 \\
EERM && 1.8624 & 0.6493 && 1.9035 & 0.6325 && 2.1187 & 0.5931&& 84.0604 & 0.4259 && 83.2518 & 0.4101  \\
DRAIN && 1.9798 & 0.6132 && 1.9969 & 0.6009 && 2.2100 &0.5741 && 91.4561 & 0.3987 && 104.3200 & 0.4085  \\\hline\hline
  HyperGNN-1 && 2.7566 & 0.2574 && 2.8543 & 0.2494 && 2.8863 & 0.2501&& 129.6152 & 0.3566 && 101.0478 & 0.4095  \\
  HyperGNN-2 && 2.9383 & 0.2654 && 3.0230 & 0.2565 && 3.0467 &0.2574 && 280.5912 & 0.3557 && 400.0514 & 0.3125  \\
  HyperGNN && 1.9720 & 0.6144 && 2.2040 & 0.5700 && 2.1930 & 0.5799&& 69.3157 & 0.4405 && 70.0319 & 0.4299  \\\hline\hline
    MultiHyperGNN-MLP && 2.8958 & 0.2608  &&3.4251  & 0.2736 && 3.6073 & 0.2814 && 104.0525 & 0.3764 && 81.9324 & 0.4122  \\
  MultiHyperGNN-S && 2.0023 & 0.6492 && 2.2420 & 0.6018 && 2.2723 &0.6153 && 89.1604 & 0.4027 && 75.6518 & 0.4151  \\
  MultiHyperGNN-GCN && 1.8023 & 0.6511 && 1.9426 & 0.6340 && 1.9539 &0.6337 && 89.1321 & 0.4395 && 68.7137 & 0.4216  \\
  MultiHyperGNN-GIN && \textbf{1.7101} &\textbf{0.6654} &&\textbf{1.8913}  &\textbf{0.6450}  &&1.9046  &0.6455 && \textbf{43.5142} & \textbf{0.5155} && \textbf{49.0168} & \textbf{0.4878}  \\
  MultiHyperGNN-GAT && 1.7695  & 0.6583 && 1.9107 & 0.6347 && \textbf{1.8951} & \textbf{0.6470}&& 54.2913 & 0.4937 &&60.1922  & 0.4561  \\\hline
\end{tabular}
\end{adjustbox}
\label{tab:pred}
\end{table*}

\begin{table*}[!tb]
\caption{Evaluation on domain-generalization accuracy. Each time we train to predict two output modes and treat another mode as the unseen target mode in testing process.}
\centering
\begin{adjustbox}{max width=0.58\textwidth}
\begin{tabular}{c c cc c cc c cc c cc}\hline
\multirow{2}{*}{Model} && \multicolumn{2}{c}{Genes-L} && \multicolumn{2}{c}{Genes-LG} && \multicolumn{2}{c}{Genes-S}  \\\cline{3-4} \cline{6-7} \cline{9-10}
&& MSE & PCC && MSE & PCC && MSE & PCC \\ \hline\hline
 ED-GNN && 2.2387 & 0.4752 &&2.0573 &0.5229 &&2.0425  & 0.5511  \\
IN && 2.1017 & 0.5312 && 2.1539 &0.5249 && 2.3746 & 0.4795  \\
EERM && 2.2148 & 0.5193 && 2.3536 & 0.4583&& 2.5792 &  0.4669  \\
DRAIN && 2.8155 & 0.5123 && 3.2461 &0.4016 && 3.2777 & 0.4230  \\\hline\hline 
  HyperGNN-1 && 3.7586 & 0.2359 && 3.3152 & 0.2614 && 3.3011 & 0.2537  \\
  HyperGNN-2 && 3.1516 & 0.2338 &&3.3064 &0.2572 &&3.5869  & 0.2629  \\
  HyperGNN && 1.9025 & 0.6003 && 2.0471 & 0.6427&& 1.9913 & 0.6236  \\\hline\hline
  MultiHyperGNN-MLP && 3.0812 & 0.2150 && 3.1519 & 0.2963&&3.6322  & 0.3049  \\
  MultiHyperGNN-GCN && 1.8513 & 0.6495 && 2.0086 &0.6410 &&1.9965  & 0.6127  \\
  MultiHyperGNN-GIN &&\textbf{1.8005}  &\textbf{0.6600}  &&\textbf{1.9852}  &\textbf{0.6479} && 1.9031 & \textbf{0.6471}  \\
  MultiHyperGNN-GAT &&1.8069  &0.6562  &&2.0123  &0.6455 && \textbf{1.8921} & 0.6425  \\\hline
\end{tabular}
\end{adjustbox}
\label{tab:dg}
\end{table*}

\subsection{Evaluation metrics}
We evaluated the model performance both quantitatively and qualitatively. For quantitative evaluation, we measured prediction accuracy based on Mean Squared Error (MSE) and Pearson Correlation Coefficients (PCC). To evaluate the efficiency, we theoretically analyzed the space complexity of MultiHyperGNN and other models. For qualitative evaluation, we visualized the distribution between predicted and ground-truth node attributes in unseen modes during the testing process. 
\subsection{Competing models and ablation studies}
We employed five competing models to compare with MultiHyperGNN regarding prediction and domain generalization: (1) \textbf{ED-GNN}. We modified MultiHyperGNN to a naive encoder-decoder-based graph transformation model by directly training the encoder and the decoder for each mode combination. A single model is trained for all mode combinations; (2) \textbf{Multi-Head Model (MHM)}. Following \citep{vandenhende2021multi}, we modified ED-GNN into a multi-task learning framework by simultaneously training multiple decoders with the same encoder. This model can only be used for the prediction purpose instead of domain generalization since each decoder deals with a specific output mode; (3) \textbf{Interaction Networks (IN)~\citep{battaglia2016interaction}}. IN models the interactions and dynamics of nodes in the graph for node-level graph transformation. Particularly, IN uses only fixed graph topology from the input mode; (4) \textbf{Explore-to-Extrapolate Risk Minimization (EERM)~\citep{wu2022handling}}. EERM employs $p$ context explorers that undergo adversarial training to maximize the variance of risks across multiple virtual environments. This design enables the model to extrapolate from a single observed environment; (4) \textbf{DRAIN~\citep{bai2022temporal}}. DRAIN utilizes a recurrent graph generation approach to generate dynamic graph-structured neural networks using hypernetworks trained on various time points. This framework can capture the temporal drift of both model parameters and data distributions, enabling it to make future predictions. In addition, we modified MultiHyperGNN to evaluate four different aspects: (1) \textbf{HyperGNN}. HyperGNN is a simpler version of MultiHyperGNN by predicting multiple output modes from one single input mode. In this case, only $\beta_d$ was trained; (2) \textbf{HyperGNN-1}. To explore whether a single MLP prediction layer can predict for all output modes, for HyperGNN-1, we will not produce MLP layers but only produce GAT layers by hypernetwork; (3) \textbf{HyperGNN-2}. In our experimental setting the meta information is composed of mode types (one-hot vector) and other mode-related features. For HyperGNN-2, we reduced the meta information by only feeding the mode type to hypernetworks; (4) \textbf{MultiHyperGNN-S}. Graph transformation from multiple input modes is expected to power the prediction by aggregating from these input modes. To validate this assumption, during the testing process of MultiHyperGNN, we will not use only a single input mode as the input data.
\subsection{Quantitative evaluation}
\subsubsection{Prediction accuracy}
\label{sec:pred}
\setlength{\intextsep}{0mm}
\setlength{\columnsep}{2mm}
\begin{wrapfigure}{r}{0.35\textwidth}
  \begin{center}
    \includegraphics[width=0.35\textwidth]{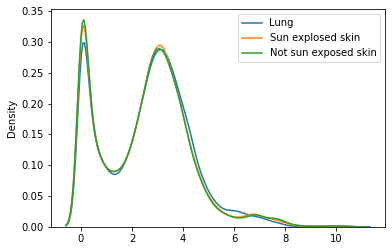}
    \label{fig:distest}
  \end{center}
  \caption{Density plot to visualize the gene expression in lung, sun-exposed skin and not-sun-exposed skin in the testing data.}
\end{wrapfigure}
On \textbf{Genes}, we trained MultiHyperGNN that predicts gene expression in lung (Genes-L), sun-exposed skin (Genes-LG) and not-sun-exposed skin (Genes-S) using gene expression of whole blood and muscle skeletal. For MultiHyperGNN-S, during testing, we used gene expression from whole blood as a single input mode. We trained HyperGNN and its variations (i.e., HyperGNN-1, HyperGNN-2) to predict in the same output tissues but only from whole blood as the single input mode. EERM and DRAIN were also trained from one single mode. For the dataset \textbf{Climate}, we trained MultiHyperGNN from the air temperature in the early morning and late morning to predict air temperature in the afternoon (T-Afternoon) and at night (T-Night). To train HyperGNN and its variations, we predicted T-Afternoon and T-Night from only late night. EERM and DRAIN were also trained from one single mode. For ED-GNN and IN, we trained them on all input-output mode combinations. To train MHM, we followed the same training strategy as we trained HyperGNN. 

As shown in Table~\ref{tab:pred}, MultiHyperGNN achieves superior performance on both datasets. The MSE of MultiHyperGNN-GIN is 0.1262 (6.43$\%$) smaller than the second best model, EERM, by average. The PCC of MultiHyperGNN-GIN is 0.0270 (4.32$\%$) higher than the second best model, EERM, by average. This is expected since MultiHyperGNN involves two input modes so that it is more expressive than EERM. HyperGNN, MultiHyperGNN-S and other competing models have comparable results since they all predict from a single input mode. The performance of HyperGNN-1 is worse, indicating that a mode-specific prediction layer is still needed. In addition, the deployment of MultiHyperGNN hinges upon the accessibility of mode-specific meta-information. As evidenced in Table~\ref{tab:pred} and Table~\ref{tab:dg}, the utilization of HyperGNN-2, which condenses meta-information to only the mode type, results in suboptimal prediction accuracy across almost all settings.

\subsubsection{Domain generalization}
\label{sec:dg}

We evaluated the performance of MultiHyperGNN and other models regarding domain generalization using the \textbf{Genes} dataset. To evaluate the generalization ability on a specific output mode (e.g., Genes-L), each time we trained the model to predict another two modes (e.g., Genes-LG, Genes-S) using data of whole blood and muscle skeletal as input modes. During testing time, we applied the trained model to the output mode (e.g., Genes-L) and calculated the prediction accuracy.

Based on Table~\ref{tab:dg}, MultiHyperGNNs shows consistently better performance compared with other models. Specifically, MultiHyperGNN-GIN has the MSE 0.0840 (4.24$\%$) smaller by average than the second model, HyperGNN, which has the MSE of 1.9803 by average. MultiHyperGNN-GIN has the PCC 0.0295 (4.74$\%$) higher by average than the second model, HyperGNN, which has the PCC of 0.6222 by average. The better performance of MultiHyperGNN results from the fact that MultiHyperGNN predicts from multiple input modes, which is more expressive than HyperGNN that only achieves single-input, multi-output mode prediction. The superior performance of MultiHyperGNN and HyperGNN compared with other models results from the meta information that guides hypernetworks to generalize the model to unseen domains. 
\subsubsection{Space complexity and implementation details}
\label{sec:complex}
We compared MultiHyperGNN with other models by the theoretical space complexity analysis. To train a predictive mapping that covers all mode combinations in $\cS$, ED-GNN, MHM and IN have $O(3^{N})$ encoders and decoders in total that need to be trained, leading to $O(3^{N})$ space complexity. EERM requires to train $p$ classifiers whereas DRAIN only needs a hypernetwork to produces classifiers at each time point. Therefore, EERM and DRAIN have the space complexity of $O(p)$ and $O(1)$, respectively. To train MultiHyperGNN, instead, we only need to train two hypernetworks, whose space complexity is $O(1)$ which is much smaller than competing models except DRAIN. 
\begin{figure*}[h]
\begin{center}
\includegraphics[width=0.9\textwidth]{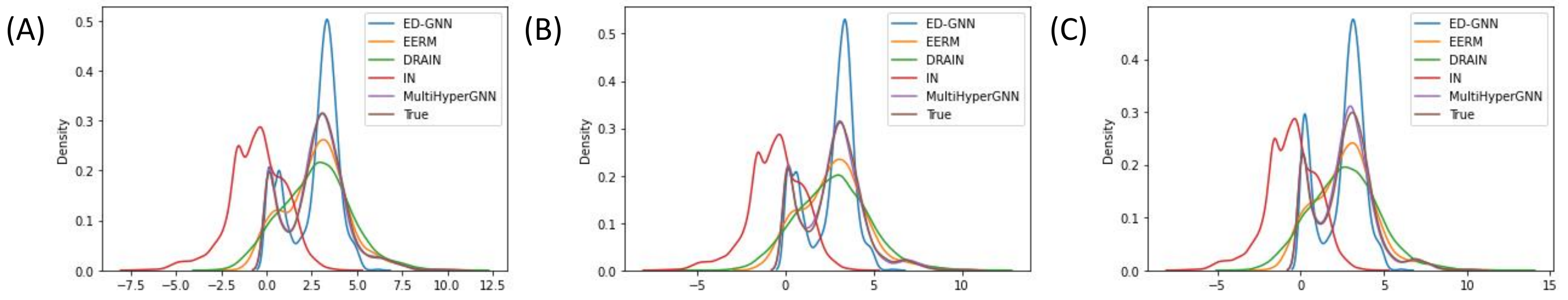}
\caption{Density plots to visualize the distribution of predicted and ground-truth gene expression in testing data, including density plot for (A) lung, (B) sun-exposed skin and (C) not-sun-exposed skin.}
\label{fig:preddist}
\end{center}
\end{figure*}
\subsection{Qualitative evaluation}
We visualized the distribution of node attributes in different modes of \textbf{Genes}. As shown in Figure~\ref{fig:distest}, in the testing data the distribution of sun-exposed skin is similar to the not-sun-exposed skin. This is reasonable since both are skin tissues and they share similar meta information. By contrast, lung is different from skin, so that its distribution is different from two skin tissues. This also confirms the necessity to design the model to handle mode similarities. We also visualized via density plots the alignment of the distribution of predicted values with the ground-truth distribution in unseen testing data (Figure~\ref{fig:preddist}) corresponding to the results in Table~\ref{tab:dg}. Based on the results in Figure~\ref{fig:preddist}, in all three human tissues, MultiHyperGNN achieves roughly the same distribution with the ground-truth distribution of the testing data, which is much better than other competing models. This is aligned with the superior prediction accuracy in domain generation of MultiHyperGNN as shown in Table~\ref{tab:dg}. 

\section{Conclusion}
\label{sec:con}
In this paper, we attempt to tackle challenges regarding domain generalization deep graph transformation. Firstly, we identify three challenges in domain generalization graph generalization. Then we propose MultiHyperGNN that includes a encoder and a decoder to respectively encode graph topologies in input and output modes. Two novel hypernetworks are designed to produce the encoder and the decoder, guided by the mode-specific meta information for domain generalization. Comprehensive experiments were conducted on real-world datasets and our model shows superior performance than competing models. Further exploration is warranted to determine the crucial components of meta-information that should be incorporated to optimize the performance of MultiHyperGNN.

\bibliographystyle{ACM-Reference-Format}
\bibliography{ref}

\end{document}


\newpage
\appendix
\section{Proof of Theorem 1}
Consider the situation that we train $f_{\gamma_{\mathcal X\rightarrow\mathcal Y}}$ (Eq. 3), parameterized by $\gamma_{\mathcal X\rightarrow \mathcal Y}$, on $\mathcal S$ s.t. $\mathcal X\times\mathcal Y\in\mathcal S$. Without loss of generalization, in the target domain, we consider predicting node attributes in mode $k$ from mode $j$, where $j\in\cX^\cT$ and $k\in\cY^\cT$. First of all, we explain that solving the following conditional likelihood is equivalent to minimizing generalization error:
\begin{eqnarray}
    m^*=\argmax_m p(\cG^{(k)}\vert \cG^{(j)}, m), \ \ j\in\cX^{\cT}, \ \ k\in\cY^\cT,
    \label{eq:suf}
\end{eqnarray}
where $m=(m^{(j)'}, m^{(k)'})$ and $m^*=(m^{(j)*}, m^{(k)*})$, $\cG^{(j)}=\{G_1^{(j)}, G_2^{(j)}, ..., G_n^{(j)}\}$ and $\cG^{(k)}=\{G_1^{(k)}, G_2^{(k)}, ..., G_n^{(k)}\}$. Then, we prove that $m^*$ that satisfies Eq.~\ref{eq:suf} should be sufficient meta information of mode $j\in\cX^\cT$ and $k\in\cY^\cT$.

\subsection{Solving Eq.~\ref{eq:suf} is equivalent to minimizing generalization error}
Since $X_i^{(k)}=f_{\hat \gamma_{\cX\rightarrow\cY}}(A^{(k)}, \{G_i^{(j)}, m^{(j)}\}_{j\in\cX^\cT}, m^{(k)};\hat \gamma_{\cX\rightarrow\cY})+\bm{\epsilon}_i=\hat X_i^{(k)} + \bm\epsilon_i$ where $\hat \gamma_{\cX\rightarrow\cY}$ is learned and fixed, considering $m=\{m^{(j)'},m^{(k)'}\}$ as the random variable, we have $\hat X_i^{(k)}-X_i^{(k)}=\bm\epsilon_i\sim\mathcal{N}(\bm{0}, \bm\sigma^2)$. As a result, take a logarithm of $p(\cG^{(k)}\vert \cG^{(j)}, m)$, we have:
\begin{eqnarray}
    \log p(\cG^{(k)}\vert \cG^{(j)}, m)&=&\log p(\{X_i^{(k)}\}_{i=1}^n, A^{(k)}\vert \{G_i^{(j)}\}_{i=1}^n, m)=\nonumber \\
    &=&\log p(\{X_i^{(k)}\}\vert A^{(k)}, \{G_i^{(j)}\}_{i=1}^n, m) \nonumber \\
    &+&\log p(A^{(k)}\vert \{G_i^{(j)}\}_{i=1}^n, m)
\end{eqnarray}
Since $A^{(k)}$ is known, then maximizing Eq.~\ref{eq:suf} is equivalent to maximizing the first term of the above equation. Given $n$ independent samples in the dataset:
\begin{eqnarray}
    m^* = \argmax_m\log p(\{X_i^{(k)}\}\vert A^{(k)}, \{G_i^{(j)}\}_{i=1}^n, m)=\argmax_m -\sum_{i=1}^n\|\frac{\hat X_i^{(k)}-X_i^{(k)}}{\bm\sigma}\|_2^2 + C,
\end{eqnarray}
where $C$ is a constant. The above objective is equivalent to minimizing the generalization error $\|\bm\epsilon_i\|_2^2=\|\hat X_i^{(k)}-X_i^{(k)}\|_2^2$.

\subsection{The sufficient meta information of $j\in\cX^\cT$ and $k\in\cY^\cT$ satisfies Eq.~\ref{eq:suf}}
Based on the Bayes' theorem, we have $p(\cG^{(j)},\cG^{(k)}\vert ,m^{(j)'}, m^{(k)'})=p(\cG^{(k)}\vert\cG^{(j)},m^{(j)'})\cdot p(\cG^{(j)}\vert m^{(j)'})$. If $m^{(j)'}$ and $m^{(k)'}$ are sufficient meta information $m^{(j)}$ and $m^{(k)}$. Therefore, $p(\cG^{(j)}\vert m^{(j)'})=1$ so that we have $p(\cG^{(j)},\cG^{(k)}\vert ,m^{(j)'}, m^{(k)'})=p(\cG^{(k)}\vert\cG^{(j)},m^{(j)'})$. Also, we have $m^{(k)}=\argmax_{m} p(\cG^{(k)}\vert \cG^{(j)}, m)$, equivalently, $p(\cG^{(j)},\cG^{(k)}\vert ,m^{(j)'}, m^{(k)'})$ is maximized by $m^{(k)}$. In conclusion, the sufficient meta information of $j\in\cX^\cT$ and $k\in\cY^\cT$ satisfies Eq.~\ref{eq:suf}.

\section{Statistics of datasets in experiments}
\textbf{Genes}. We used gene expression data collected and curated by the Genotype-Tissue Expression (GTEx) Consortium~\citep{lonsdale2013genotype}. Specifically, processed gene expression data derived from bulk RNA-seq experiments on five tissues, whole blood (WB), lung (L), muscle skeletal (MS), sun-exposed skin (lower leg, LG), not-sun-exposed skin (suprapubic, S) were used. For quality control purpose, we first removed samples with no data in any of the five tissues types. Then we removed genes with low expression level (total number of mapped reads less than 2 across all samples). For the remaining data, for each tissue, we performed weighted correlation network analysis (WGCNA)~\citep{langfelder2008wgcna} with the cutoff $\rho$ on the correlation coefficients to construct the co-expression network with the expression value as the node attribute. The meta information that characterizes these five tissues includes tissue type (lung, muscle, skin), location (trunk, leg, arm), structure (dense, rigid, spongy), function (movement, protection, gas exchange) and cell types (alveoli and bronchioles, cylindrical muscle fibers, epithelial cells).

\textbf{Climate}. We used the Goddard Earth Observing System Composition Forecasting (GEOS-CF) hourly historical meteorological data\footnote{\url{https://gmao.gsfc.nasa.gov/weather_prediction/GEOS-CF/}} across the contiguous United States from 2019-2021. The GEOS-CF meteorological data was assimilated from a variety of conventional and satellite-driven data sources. The detailed assimilation approaches can be found at the website\footnote{\url{https://ntrs.nasa.gov/citations/20120011955}}. We further calculated the surface air temperature (T) for each state capital by averaging data from all the GEOS-CF pixels that fall within a given capital city plus a 10 km buffer region. Specifically, we splitted 24 hours of a day into four time periods as four modes: early morning (0:00AM-6:00AM), late morning (6:00AM-12:00PM), afternoon (12:00PM-18:00PM) and night (18:00PM-0:00AM), and calculated the mean value of the air temperature in each period. To construct the network in each domain, we used cities as graph nodes and air temperature in each city as the node attribute. Then in each time period, we calculated the correlation of the air temperature between two cities within three years. If the correlation is greater than $\rho$ then there is an edge connecting two cities on the graph. We used the time period indicator (four-element, one-hot vector to indicate early morning, late morning, afternoon and night) and various time stamps when the data was collected as the meta information.

The statistics of these two datasets are in Table~\ref{tab:data}.

\section{Implementation details}
All experiments were conducted by Python 3.9 on the 64-bit machine with an NVIDIA GPU, NVIDIA GeForce RTX 3090. In practice we use six-layer MLPs with the hidden dimension as 7200 to model $\gamma_e$ and $\gamma_d$. The source encoder and the target decoder contain two-layer GNNs (with four heads if GATs) and the hidden dimension of 256. The prediction layer is composed of five-layer MLPs with the hidden dimension of 512.


\begin{table}[!tb]
\caption{Statistics of employed datasets, \textbf{Genes} and \textbf{Climate} ($|\mathcal{D}|$ number of domains contained in the dataset; $|\mathcal{V}|$ is the size of the graph in the dataset; $|\mathcal{E}|_{avg}$ is the average number of edges of the graph in the dataset by domain; $n$ is the number of samples in each dataset).}
\centering
\begin{adjustbox}{max width=0.6\textwidth}
\begin{tabular}{c c c c c}
\hline
Model && Genes && Climate  \\ \hline
   $|\mathcal{D}|$  && 5 && 4 \\ \hline
   $|\mathcal{V}|$  && 2,398 && 48\\ \hline
   $|\mathcal{E}|_{avg}$  && 28477/1174/1020/7463/272  && 293/266/336/272    \\ \hline
   $n$  && 205 && 1,095 \\ \hline
\end{tabular}
\end{adjustbox}
\label{tab:data}
\end{table}